# Context-Based Word Acquisition for Situated Dialogue in a Virtual World

**Shaolin Qu**                                                    QUSHAOLI@CSE.MSU.EDU
**Joyce Y. Chai**                                                      JCHAI@CSE.MSU.EDU
*Department of Computer Science and Engineering*
*Michigan State University*
*East Lansing, MI 48824 USA*

## Abstract

To tackle the vocabulary problem in conversational systems, previous work has applied unsupervised learning approaches on co-occurring speech and eye gaze during interaction to automatically acquire new words. Although these approaches have shown promise, several issues related to human language behavior and human-machine conversation have not been addressed. First, psycholinguistic studies have shown certain temporal regularities between human eye movement and language production. While these regularities can potentially guide the acquisition process, they have not been incorporated in the previous unsupervised approaches. Second, conversational systems generally have an existing knowledge base about the domain and vocabulary. While the existing knowledge can potentially help bootstrap and constrain the acquired new words, it has not been incorporated in the previous models. Third, eye gaze could serve different functions in human-machine conversation. Some gaze streams may not be closely coupled with speech stream, and thus are potentially detrimental to word acquisition. Automated recognition of closely-coupled speech-gaze streams based on conversation context is important. To address these issues, we developed new approaches that incorporate user language behavior, domain knowledge, and conversation context in word acquisition. We evaluated these approaches in the context of situated dialogue in a virtual world. Our experimental results have shown that incorporating the above three types of contextual information significantly improves word acquisition performance.

## 1. Introduction

One major bottleneck in human machine conversation is robust language interpretation. When the encountered vocabulary is outside of the system's knowledge, the system tends to fail. As conversational interfaces have become increasingly important in many applications such as remote interaction with robots (Lemon, Gruenstein, & Peters, 2002; Fong & Nourbakhsh, 2005) and automated training and education (Traum & Rickel, 2002), the ability to automatically acquire new words during online conversation becomes essential. Different from traditional telephony-based spoken dialogue systems, in conversational interfaces users can look at a graphic display or a virtual world while interacting with artificial agents using natural speech. This unique setting provides an opportunity for automated vocabulary acquisition. During interaction, users' visual perception (e.g., as indicated by eye gaze) provides a potential channel for the system to automatically learn new words.

The idea, as shown in previous work (Yu & Ballard, 2004; Liu, Chai, & Jin, 2007), is that the parallel data of visual perception and spoken utterances can be used by unsupervised approaches to automatically identify the mappings between words and visual entities and thus acquire new words. While previous approaches provide a promising direction, they mainly rely on the co-occurrence between words and visual entities in a completely unsupervised manner. However, in human machine conversation, there are different types of extra information related to human language behaviors and characteristics of conversation systems. Although this extra information can potentially provide supervision to guide the word acquisition process and improve performance, it has not been systematically explored in previous work.

First, a large body of psycholinguistic studies have shown that eye gaze is tightly linked to human language processing. This is evident in both language comprehension (Tanenhaus, Spivey-Knowiton, Eberhard, & Sedivy, 1995; Eberhard, Spivey-Knowiton, Sedivy, & Tanenhaus, 1995; Allopenna, Magnuson, & Tanenhaus, 1998; Dahan & Tanenhaus, 2005; Spivey, Tanenhaus, Eberhard, & Sedivy, 2002) and language production (Meyer, Sleiderink, & Levelt, 1998; Rayner, 1998; Griffin & Bock, 2000; Bock, Irwin, Davidson, & Leveltb, 2003; Brown-Schmidt & Tanenhaus, 2006; Griffin, 2001). Specifically in human language production, which is directly relevant to automated computer interpretation of human language, studies have found significant temporal regularities between the mentioned objects and the corresponding words (Meyer et al., 1998; Rayner, 1998; Griffin & Bock, 2000). In object naming tasks, the onset of a word begins approximately one second after a speaker has looked at the corresponding visual referent (Griffin, 2004), and gazes are longer the more difficult the name of the referent is to retrieve (Meyer et al., 1998; Griffin, 2001). About 100-300 ms after the articulation of the object name begins, the eyes move to the next object relevant to the task (Meyer et al., 1998). All these findings suggest that eyes move to the mentioned objects before the corresponding words are uttered. Although this language behavior can be used to constrain the mapping between words and visual objects, it has not been incorporated in the previous approaches.

Second, practical conversational systems always have existing knowledge about their application domains and vocabularies. This knowledge base is acquired at their development time: either authored by the domain experts or automatically learned from available data. Although the existing knowledge can be rather limited and is desired to be enhanced automatically online (e.g., through automated vocabulary acquisition), it provides important information about the structure of the domain and the existing vocabularies, which can further bootstrap and constrain new word acquisition. This type of domain knowledge has not been utilized in previous approaches.

Third, although psycholinguistic studies provide us with a sound empirical basis for assuming that eye movements are predictive of speech, the gaze behavior in an interactive setting can be much more complex. There are different types of eye movements (Kahneman, 1973). The naturally occurring eye gaze during speech production may serve different functions, for example, to engage in the conversation or to manage turn taking (Nakano, Reinstein, Stocky, & Cassell, 2003). Furthermore, while interacting with a graphic display, a user could be talking about objects that were previously seen on the display or something completely unrelated to any object the user is looking at. Therefore using all the speech-gaze pairs for word acquisition can be detrimental. The type of gaze that is mostly useful for word acquisition is the kind that reflects the underlying attention and tightly links to the content of the co-occurring speech. Thus, automatically recognizing closely coupled speech and gaze streams during online conversation for word acquisition is important. However, it has not been examined in previous work.

To address the above three issues, we have developed new approaches to automatic word acquisition that (1) incorporate findings on user language behavior from psycholinguistic studies, in particular the temporal alignment between spoken words and eye gaze; (2) utilize the existing domain knowledge, and (3) automatically identify closely-coupled speech and gaze streams based on conversation context. We evaluated these approaches in the context of situated dialogue in a virtual world. Our experimental results have shown that incorporating the above three types of contextual information significantly improves word acquisition performance. Our simulation studies further demonstrate the effect of automatic online word acquisition on improving language understanding in human-machine conversation.

In the following sections, we first introduce the domain and data collection in our investigation. We then describe the enhanced models for word acquisition that incorporate additional contextual information (e.g., language behavior between spoken words and eye gaze, domain knowledge, conversation context). Finally, we present the empirical evaluation of our enhanced models and demonstrate the effect of online word acquisition on spoken language understanding during human-machine conversation.

## 2. Related Work

Our work is motivated by previous work on grounded language acquisition and eye gaze in multimodal human-computer interaction.

Grounded language acquisition is to learn the meaning of language by connecting the language to the perception of the world. Language acquisition by grounding words to the visual perceptions of objects has been studied in various language learning systems. For example, given speech paired with video images of single objects, mutual information between audio and visual signals was used to learn words by associating acoustic phoneme sequences with the visual prototypes (e.g., color, size, shape) of objects (Roy & Pentland, 2002; Roy, 2002). Generative models were developed to learn words by associating words with image regions given parallel data of pictures and description text (Barnard, Duygulu, de Freitas, Forsyth, Blei, & Jordan, 2003). Given pairs of spoken instructions containing object names and the corresponding objects, an utterance-object joint probability model was used to learn object names by identifying object name phonemes and associating them with the objects (Taguchi, Iwahashi, Nose, Funakoshi, & Nakano, 2009). Given sequences of utterances paired with scene representations, an incremental translation model was developed to learn word meaning by associating words with the semantic representations of the referents in the scene (Fazly, Alishahi, & Stevenson, 2008). In addition to grounding individual words, previous work has also investigated grounding phrases (referring expressions) to visual objects through semantic decomposition, for example using context free grammar that connects linguistic structures with underlying visual properties (Gorniak & Roy, 2004).

Besides visual objects, approaches have also been developed to ground words to meaning representations of events. For example, event logic was applied to ground verbs to motion events that were represented by force dynamics encoding the support, contact, and attachment relations between objects in video images (Siskind, 2001). In a video game domain, a translation model was used to ground words to the semantic roles of user actions (Fleischman & Roy, 2005). In a simulated Robocup soccer domain, given textual game commentaries paired with the symbolic descriptions of game events, approaches based on statistical parsing and learning were developed to ground the commentary text to game events (Chen & Mooney, 2008). In a less restricted data setting, generative models were developed to simultaneously segment the text into utterances and map the utterances to meaning representations of event states (Liang, Jordan, & Klein, 2009). Different from the above previous work, in our work, the visual perception is indicated by eye gaze. Eye gaze, on one hand, is indicative of human attention, which provides opportunities to link language and perception; on the other hand, is an implicit and subconscious input, which could bring additional challenge in word acquisition.

Eye gaze has long been explored in human-computer interaction for direct manipulation interfaces as a pointing device (Jacob, 1991; Wang, 1995; Zhai, Morimoto, & Ihde, 1999). Eye gaze as a modality in multimodal interaction goes beyond the function of pointing. In different speech and eye gaze systems, eye gaze has been explored for the purpose of mutual disambiguation (Tanaka, 1999; Zhang, Imamiya, Go, & Mao, 2004), as a complement to the speech channel for reference resolution (Campana, Baldridge, Dowding, Hockey, Remington, & Stone, 2001; Kaur, Termaine, Huang, Wilder, Gacovski, Flippo, & Mantravadi, 2003; Prasov & Chai, 2008; Byron, Mampilly, Sharma, & Xu, 2005) and speech recognition (Cooke, 2006; Qu & Chai, 2007), and for managing human-computer dialogue (Qvarfordt & Zhai, 2005).

Eye gaze has been explored recently for word acquisition. For example, Yu and Ballard (2004) proposed an embodied multimodal learning interface for word acquisition, especially through eye movement. In this work, given speech paired with eye gaze information and video images, a translation model was applied to acquire words by associating acoustic phone sequences with visual representations of objects and actions. This work has inspired our research and is mostly related to our effort here. The difference between our work and the work by Yu and Ballard lies in two aspects. First, the learning environment is different. While Yu and Ballard focuses on the narrative descriptions of actions (e.g., making a sandwich, pouring some drinks, etc.) from human subjects, our focus is on interactive conversation. In conversation, a human participant can take both a speaker role and an addressee role. This represents a new scenario where word acquisition based on eye movement may have new implications. Second, in the work by Yu and Ballard, the IBM Translation Model 1 was applied to word acquisition. In our work, we further incorporate other types of information such as user language behavior, domain knowledge, and conversation context in the translation models.

In our previous work, we have experimented the application of IBM Translation Model 1 in vocabulary acquisition through gaze modeling in a conversation setting (Liu et al., 2007). We have reported our initial investigation on incorporating temporal information and domain knowledge in translation models (Qu & Chai, 2008) as well as automatically identifying closely coupled speech and gaze streams (Qu & Chai, 2009). This paper extends our previous work and provides a comprehensive evaluation on incorporating knowledge and interactivity in word acquisition in a much richer application domain. We further examine how word acquisition is affected by automated speech recognition and what is the effect of online word acquisition on language understanding in human-machine conversation.

## 3. Domain and Data

To facilitate our work on word acquisition, we collected data based on situated dialogue in a virtual world. This data set is different from the data set used in our previous investigation (Qu & Chai, 2008). The difference lies in two aspects: 1) this dataset was collected during mixed initiative human-machine conversation whereas the data in our previous investigation was based only on question and answering; 2) user studies in this work were conducted in the context of situated dialogue, where human users are immersed in a complex virtual world and can move around in the virtual environment.

### 3.1 Domain

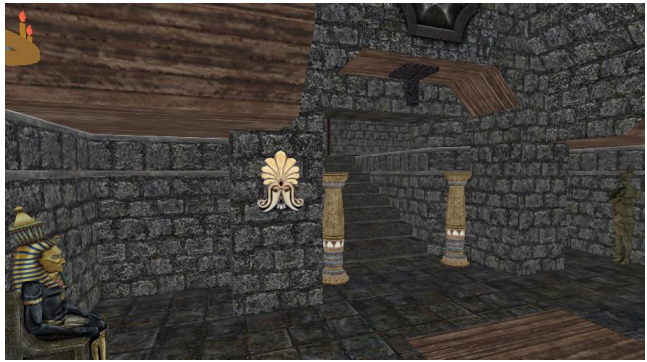

Figure 1: Treasure hunting domain

The application of our virtual world is for treasure hunting. A human user is immersed in a virtual world (i.e., a castle) [1] and is able to move around to look for hidden treasures. The castle contains 115 3D objects. The user needs to consult with a remote "expert" (i.e., the artificial agent) for advice on finding the treasures. The expert has some knowledge about the environment and the treasures but cannot see where the user is in the castle. So the user needs to describe to the expert where he is and what he sees in order to get advice on where to go and what to look for.

Figure 1 shows a snapshot of our virtual world. Figure 2 shows an excerpt of the dialog between a user and the expert in finding one of the treasures. To focus our investigation on word acquisition, we conducted wizard-of-oz studies to collect data [2]. More specifically, a set of response templates corresponding to different types of responses are predefined. At each

---

1. The virtual world was developed based on the Irrlicht game engine `http://irrlicht.sourceforge.net/`.
2. WOZ studies allow the observation of a user interacting with a fully functional system, where some or all of system functions are supplemented by a hidden human wizard. Users are led to believe they are interacting with a real system (Klemmer, Sinha, Chen, Landay, Aboobaker, & Wang, 2000). This technology, which permits the experimentally controlled study of human computer interaction behavior without exposing users to unpredicted responses, has been used widely in earlier research on speech interfaces and multimodal systems.

S: *Tell me what you see.*
U: *I see a golden tomb or sarcophagus.*
S: *Describe it.*
U: *It appears to be made of gold and there are Egyptian decorations on it.*
S: *What else do you see?*
U: *I see a carving hanging on the wall appears to be African.*
S: *Go back.*
U: *I'm back by the sarcophagus. There appears to be lava surrounding it or fire.*
S: *I couldn't understand you.*
U: *The room that the sarcophagus is in is surrounded by lava or fire.*
S: *Do you see a statue?*
U: *Should i try to open the sarcophagus?*
S: *No.*
U: *I see a statue.*
S: *Describe.*
U: *It appears to be a stone statue of an Egyptian.*
S: *What else do you see?*
U: *I see a small vase a pottery vase decorated with flowers.*
S: *The treasure is near.*
U: *Should i try to move the vase?*
S: *Yes.*
U: *I can't move the vase.*
S: *Do you see another one?*
U: *Yes, there's a wooden table with a pottery jug and a purple vase on it. Should i try to move the purple vase?*
S: *Do you see a statue?*
U: *Yes, I see a stone statue.*
S: *Do you see another one?*
U: *Yes, I see a gold and black statue of an Egyptian sitting.*
S: *Compare these objects.*
U: *The stone statue is in very bad shape, and the golden statue appears to be in good shape. It has a golden crown. It is sitting in a wooden chair.*
S: *Try moving it.*
U: *I move the head off the statue and now i see a scroll. Should i take the scroll?*
S: *Describe.*
U: *It is a white scroll with a red seal.*
S: *Take it.*
U: *I took the scroll.*
S: *Good.*

Figure 2: The dialog between a user and the system in finding one of the treasures

system turn, the system automatically determines what kind of responses (i.e., template) should be provided; for example, whether to ask for clarification, provide confirmation, or ask more information about a particular object. Such decision is made based on a distribution to simulate different types of system acceptance and rejection. This distribution can be modified according to different needs of the experiments. Based on the chosen template, the human wizard then serves as a language understanding component and fills in this template with the specific information related to the user input. The filled template is used to automatically generate natural language which is further processed by the Microsoft Text-to-Speech engine to generate speech responses. During the experiments, the user's speech was recorded, and the user's eye gaze was captured by a Tobii eye tracker.

## 3.2 Data Preprocessing

From 20 users' experiments, we collected 3709 utterances with accompanying gaze fixations. We transcribed the collected speech. The vocabulary size of the speech transcript is 1082, among which 757 are nouns/adjectives. The user's speech was also automatically recognized online by the Microsoft speech recognizer with a word error rate (WER) of 48.1% for the 1-best recognition. The vocabulary size of the 1-best speech recognition is 3041, among which 1643 are nouns/adjectives. The nouns and adjectives in the transcriptions and the recognized 1-best hypotheses were automatically identified by the Stanford Part-of-Speech Tagger (Toutanova & Manning, 2000; Toutanova, Klein, Manning, & Singer, 2003).

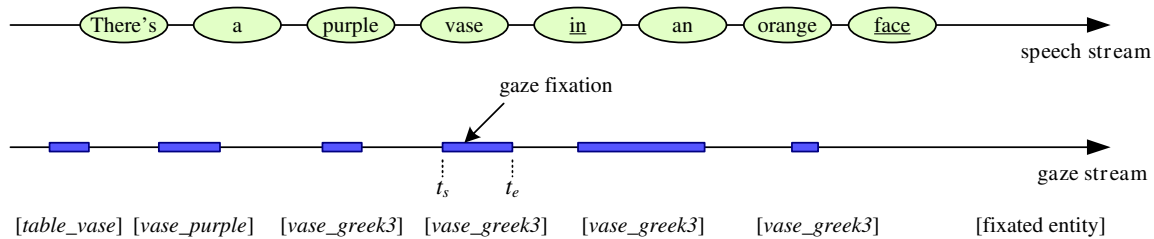

Figure 3: Accompanying gaze fixations and the 1-best recognition of a user's utterance "There's a purple vase and an orange vase." (There are two incorrectly recognized words "in" and "face" in the 1-best recognition)

The collected speech and gaze streams were automatically paired together by the system. Each time the system detected a sentence boundary of the user's speech, it paired the recognized speech with the gaze fixations that the system had been accumulating since the previously detected sentence boundary. Figure 3 shows a stream pair of a user's speech and its accompanying gaze fixations. In the speech stream, each spoken word was timestamped by the speech recognizer. In the gaze stream, each gaze fixation has a starting timestamp $t_s$ and an ending timestamp $t_e$ provided by the eye tracker. Each gaze fixation results in a fixated entity (3D object). When multiple entities are fixated by one gaze fixation due to the overlapping of the entities, the foremost one is chosen. In the gaze stream, the neighboring gaze fixations that fixate the same entity are merged.

Given the paired speech and gaze streams, we build a set of parallel word sequences and gaze fixated entity sequences $\{(\mathbf{w}, \mathbf{e})\}$ for the task of word acquisition. The word sequence $\mathbf{w}$ consists of only nouns and adjectives from the 1-best recognition of the spoken utterance. The entity sequence $\mathbf{e}$ contains the entities fixated by the gaze fixations. For the parallel speech and gaze streams shown in Figure 3, the resulting word sequence is $\mathbf{w} = $ [purple vase orange face] and the resulting entity sequence is $\mathbf{e} = [table\_vase \ vase\_purple \ vase\_greek3]$.

## 4. Translation Models for Word Acquisition

Since we are working on conversational systems where users interact with a visual scene, we consider the task of word acquisition as associating words with visual entities in the domain. Given the parallel speech and gaze fixated entities $\{(\mathbf{w}, \mathbf{e})\}$, we formulate word acquisition as a translation problem and use translation models to estimate word-entity association probabilities $p(w|e)$. The words with the highest association probabilities are chosen as acquired words for entity $e$.

### 4.1 Base Model I

Using the translation model I (Brown, Pietra, Pietra, & Mercer, 1993), where each word is equally likely to be aligned with each entity, we have

$$p(\mathbf{w}|\mathbf{e}) = \frac{1}{(l+1)^m} \prod_{j=1}^{m} \sum_{i=0}^{l} p(w_j|e_i) \tag{1}$$

where $l$ and $m$ are the lengths of the entity and word sequences respectively. We refer to this model as **Model-1**.

### 4.2 Base Model II

Using the translation model II (Brown et al., 1993), where alignments are dependent on word/entity positions and word/entity sequence lengths, we have

$$p(\mathbf{w}|\mathbf{e}) = \prod_{j=1}^{m} \sum_{i=0}^{l} p(a_j = i|j, m, l) p(w_j|e_i) \tag{2}$$

where $a_j = i$ means that $w_j$ is aligned with $e_i$. When $a_j = 0$, $w_j$ is not aligned with any entity ($e_0$ represents a *null* entity). We refer to this model as **Model-2**.

Compared to Model-1, Model-2 considers the ordering of words and entities in word acquisition. EM algorithms are used to estimate the probabilities $p(w|e)$ in the translation models.

## 5. Incorporating User Language Behavior in Word Acquisition

In Model-2, word-entity alignments are estimated from co-occurring word and entity sequences in an unsupervised way. The estimated alignments are dependent on where the words/entities appear in the word/entity sequences, not on when those words and gaze fixated entities actually occur. Motivated by the findings that users move their eyes to the mentioned object directly before speaking a word (Griffin & Bock, 2000), we make the word-entity alignments dependent on their temporal relation in a new model (referred as **Model-2t**):

$$p(\mathbf{w}|\mathbf{e}) = \prod_{j=1}^{m} \sum_{i=0}^{l} p_t(a_j = i|j, \mathbf{e}, \mathbf{w}) p(w_j|e_i) \tag{3}$$

where $p_t(a_j = i|j, \mathbf{e}, \mathbf{w})$ is the temporal alignment probability computed based on the temporal distance between entity $e_i$ and word $w_j$.

We define the temporal distance between $e_i$ and $w_j$ as

$$d(e_i, w_j) = \begin{cases} 0 & t_s(e_i) \leq t_s(w_j) \leq t_e(e_i) \\ t_e(e_i) - t_s(w_j) & t_s(w_j) > t_e(e_i) \\ t_s(e_i) - t_s(w_j) & t_s(w_j) < t_s(e_i) \end{cases} \quad (4)$$

where $t_s(w_j)$ is the starting timestamp (ms) of word $w_j$, $t_s(e_i)$ and $t_e(e_i)$ are the starting and ending timestamps (ms) of gaze fixation on entity $e_i$.

The alignment of word $w_j$ and entity $e_i$ is decided by their temporal distance $d(e_i, w_j)$. Based on the psycholinguistic finding that eye gaze happens before a spoken word, $w_j$ is not allowed to be aligned with $e_i$ when $w_j$ happens earlier than $e_i$ (i.e., $d(e_i, w_j) > 0$). When $w_j$ happens no earlier than $e_i$ (i.e., $d(e_i, w_j) \leq 0$), the closer they are, the more likely they are aligned. Specifically, the temporal alignment probability of $w_j$ and $e_i$ in each co-occurring instance $(\mathbf{w}, \mathbf{e})$ is computed as

$$p_t(a_j = i|j, \mathbf{e}, \mathbf{w}) = \begin{cases} 0 & d(e_i, w_j) > 0 \\ \dfrac{\exp[\alpha \cdot d(e_i, w_j)]}{\displaystyle\sum_{i=0}^{l} \exp[\alpha \cdot d(e_i, w_j)]} & d(e_i, w_j) \leq 0 \end{cases} \quad (5)$$

where $\alpha$ is a constant for scaling $d(e_i, w_j)$.

An EM algorithm is used to estimate probabilities $p(w|e)$ and $\alpha$ in Model-2t. It is worthwhile to mention that, findings from psycholinguistic studies have provided specific offsets in terms of how eye gaze corresponds to speech production. For example, it shows that speakers look at an object about a second before they say it, but about 100-300 ms after articulation of the object name begins, the eyes move to the next object relevant to the task (Meyer et al., 1998). Since the conversation setting in our study is much more complex than the simple settings in psycholinguistic research, we found larger variations on the offset (Liu et al., 2007) in our data. Therefore we chose not to use any offset in the alignment model here.

## 6. Incorporating Domain Knowledge in Word Acquisition

Speech-gaze temporal alignment and occurrence statistics sometimes are not sufficient to associate words to entities correctly. For example, suppose a user says "*there is a lamp on the dresser*" while looking at a lamp object on a table object. Due to their co-occurrence with the lamp object, the words *dresser* and *lamp* are both likely to be associated with the lamp object in the translation models. As a result, the word *dresser* is likely to be incorrectly acquired for the lamp object. For the same reason, the word *lamp* could be acquired incorrectly for the table object. To solve this type of association problem, the semantic knowledge about the domain and words can be helpful. For example, the knowledge that the word *lamp* is more semantically related to the object lamp can help the system avoid associating the word *dresser* to the lamp object. Specifically, we solve this type of word-entity association by utilizing the domain knowledge present in the system and external lexical semantic resources.

On one hand, each conversational system has a *domain model*, which is the knowledge representation about its domain such as the types of objects and their properties and relations, the task structures, etc. This domain model is usually acquired at the development stage before the deployment of the system. The domain model provides an important resource to enable domain reasoning and language interpretation (DeVault & Stone, 2003). On the other hand, there are available resources about domain independent lexical knowledge (e.g., WordNet, see Fellbaum, 1998). The idea is that if the domain model can be linked to the external lexical resources either manually or automatically at the development state, then the external knowledge source can be used to help constrain the acquired words.

In the following sections, we first describe our domain modeling, then define the semantic relatedness of word and entity based on domain modeling and WordNet semantic lexicon, and finally describe different ways of using the semantic relatedness of word and entity to help word acquisition.

## 6.1 Domain Modeling

We model the treasure hunting domain as shown in Figure 4. The domain model contains all domain related semantic concepts. For practical conversational systems, this domain modeling is typically acquired at the development stage either through manual authoring by the domain experts or automated learning based on annotated data. In our current work, all properties of the domain entities are represented by domain concepts. The entity properties include: *semantic type*, *color*, *size*, *shape*, and *material*. We use WordNet synsets to represent the domain concepts (i.e., synsets in the format of "word#part-of-speech#sense-id"). The "sense-id" here represents the specific WordNet sense associated with the word in representing the concept. For example, the domain concepts *SEM_PLATE* and *COLOR* of the entity *plate* are represented by the synsets "plate#n#4" and "color#n#1" in WordNet. Note that the link between domain concepts and WordNet synsets can be automatically acquired given the existing vocabularies. Here, to illustrate the idea, we simplify the problem and directly connect domain concepts to synsets.

Note that in the domain model, the domain concepts are not specific to a certain entity, they are general concepts for a certain type of entity. Multiple entities of the same type have the same properties and share the same set of domain concepts. Therefore, properties such as *color* and *size* of an entity have general concepts "color#n#1" and "size#n#1" instead of more specific concepts like "yellow#a#1" and "big#a#1", so their concepts can be shared by other entities of the same type, but with different colors and sizes.

## 6.2 Semantic Relatedness of Word and Entity

We compute the semantic relatedness of a word $w$ and an entity $e$ based on the semantic similarity between $w$ and the properties of $e$ using the domain model as a bridge. Specifically, semantic relatedness $SR(e, w)$ is defined as

$$SR(e, w) = \max_{i,j} sim(s(c_e^i), s_j(w)) \qquad (6)$$

where $c_e^i$ is the $i$-th property of entity $e$, $s(c_e^i)$ is the synset of property $c_e^i$ as in domain model, $s_j(w)$ is the $j$-th synset of word $w$ as defined in WordNet, and $sim(\cdot, \cdot)$ is the similarity score of two synsets.

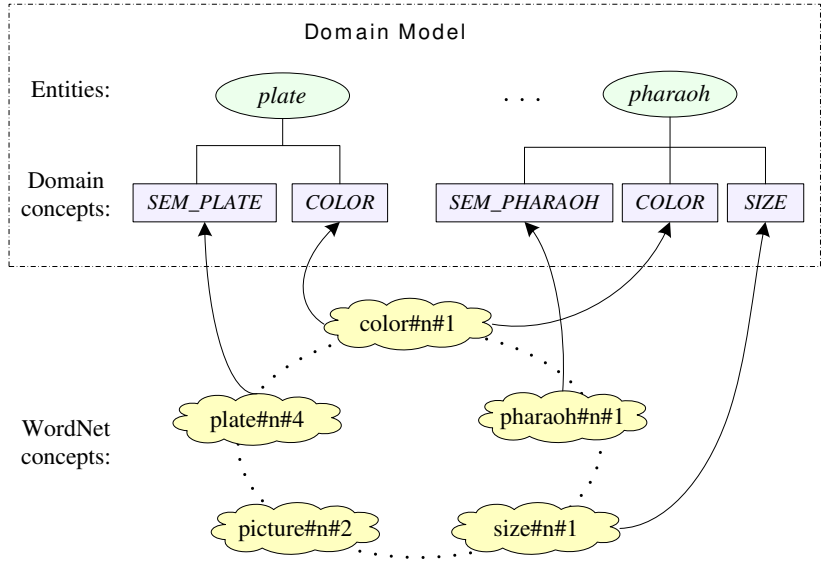

Figure 4: Domain model with domain concepts represented by WordNet synsets

We computed the similarity score of two synsets based on the path length between them. The similarity score is inversely proportional to the number of nodes along the shortest path between the synsets as defined in WordNet. When the two synsets are the same, they have the maximal similarity score of 1. The WordNet-Similarity tool (Pedersen, Patwardhan, & Michelizzi, 2004) was used for the synset similarity computation.

## 6.3 Word Acquisition with Word-Entity Semantic Relatedness

We can use the semantic relatedness of word and entity to help the system acquire semantically compatible words for each entity, and therefore improve word acquisition performance. The semantic relatedness can be applied for word acquisition in two ways: post-process the learned word-entity association probabilities by rescoring them with semantic relatedness, or directly affect the learning of word-entity associations by constraining the alignment of word and entity in the translation models.

### 6.3.1 Rescoring with Semantic Relatedness

In the acquired word list for an entity $e_i$, each word $w_j$ has an association probability $p(w_j|e_i)$ that is learned from a translation model. We use the semantic relatedness $SR(e_i, w_j)$ to redistribute the probability mass for each $w_j$. The new association probability is given by:

$$p'(w_j|e_i) = \frac{p(w_j|e_i)SR(e_i,w_j)}{\sum_j p(w_j|e_i)SR(e_i,w_j)} \tag{7}$$

### 6.3.2 Semantic Alignment Constraint in Translation Model

When used to constrain the word-entity alignment in the translation model, semantic relatedness can be used alone or used together with speech-gaze temporal information to decide the alignment probability of word and entity (Qu & Chai, 2008).

- Using only semantic relatedness to constrain word-entity alignments in **Model-2s**, we have

$$p(\mathbf{w}|\mathbf{e}) = \prod_{j=1}^{m} \sum_{i=0}^{l} p_s(a_j = i|j, \mathbf{e}, \mathbf{w}) p(w_j|e_i) \qquad (8)$$

where $p_s(a_j = i|j, \mathbf{e}, \mathbf{w})$ is the alignment probability based on semantic relatedness,

$$p_s(a_j = i|j, \mathbf{e}, \mathbf{w}) = \frac{SR(e_i, w_j)}{\sum_i SR(e_i, w_j)} \qquad (9)$$

- Using semantic relatedness and temporal information to constrain word-entity alignments in **Model-2ts**, we have

$$p(\mathbf{w}|\mathbf{e}) = \prod_{j=1}^{m} \sum_{i=0}^{l} p_{ts}(a_j = i|j, \mathbf{e}, \mathbf{w}) p(w_j|e_i) \qquad (10)$$

where $p_{ts}(a_j = i|j, \mathbf{e}, \mathbf{w})$ is the alignment probability that is decided by both temporal relation and semantic relatedness of $e_i$ and $w_j$,

$$p_{ts}(a_j = i|j, \mathbf{e}, \mathbf{w}) = \frac{p_s(a_j = i|j, \mathbf{e}, \mathbf{w}) p_t(a_j = i|j, \mathbf{e}, \mathbf{w})}{\sum_i p_s(a_j = i|j, \mathbf{e}, \mathbf{w}) p_t(a_j = i|j, \mathbf{e}, \mathbf{w})} \qquad (11)$$

where $p_s(a_j = i|j, \mathbf{e}, \mathbf{w})$ is the semantic alignment probability in Equation (9), and $p_t(a_j = i|j, \mathbf{e}, \mathbf{w})$ is the temporal alignment probability given in Equation (5).

EM algorithms are used to estimate $p(w|e)$ in Model-2s and Model-2ts.

## 7. Incorporating Conversation Context in Word Acquisition

As mentioned earlier, not all speech-gaze pairs are useful for word acquisition. In a speech-gaze pair, if the speech does not have any word that relates to any of the gaze fixated entities, this instance only adds noise to word acquisition. Therefore, we should identify the closely coupled speech-gaze pairs and only use them for word acquisition.

In this section, we first describe the feature extraction based on conversation interactivity, then describe the use of a logistic regression classifier to predict whether a speech-gaze pair is a **closely coupled speech-gaze instance** – an instance where at least one noun or adjective in the speech stream is referring to some gaze fixated entity in the gaze stream. For the training of the classifier for speech-gaze prediction, we manually labeled each instance whether it is a closely coupled speech-gaze instance based on the speech transcript and gaze fixations.

### 7.1 Features Extraction

For a parallel speech-gaze instance, the following sets of features are automatically extracted.

#### 7.1.1 Speech Features (S-Feat)

Let $c_w$ be the count of nouns and adjectives in the utterance, and $l_s$ be the temporal length of the speech. The following features are extracted from speech:

- $c_w$ – count of nouns and adjectives.
  More nouns and adjectives are expected in the user's utterance describing entities.

- $c_w/l_s$ – normalized noun/adjective count.
  The effect of speech length $l_s$ on $c_w$ is considered.

#### 7.1.2 Gaze Features (G-Feat)

For each fixated entity $e_i$, let $l_e^i$ be its fixation temporal length. Note that several gaze fixations may have the same fixated entity, $l_e^i$ is the total length of all the gaze fixations that fixate on entity $e_i$. We extract the following features from gaze stream:

- $c_e$ – count of different gaze fixated entities.
  Less fixated entities are expected when the user is describing entities while looking at them.

- $c_e/l_s$ – normalized entity count.
  The effect of speech temporal length $l_s$ on $c_e$ is considered.

- $\max_i(l_e^i)$ – maximal fixation length.
  At least one fixated entity's fixation is expected to be long enough when the user is describing entities while looking at them.

- $mean(l_e^i)$ – average fixation length.
  The average gaze fixation length is expected to be longer when the user is describing entities while looking at them.

- $var(l_e^i)$ – variance of fixation lengths.
  The variance of the fixation lengths is expected to be smaller when the user is describing entities while looking at them.

The number of gaze fixated entities is not only decided by the user's eye gaze, it is also affected by the visual scene. Let $c_e^s$ be the count of all the entities that have been visible during the length of the gaze stream. We also extract the following scene related feature:

- $c_e/c_e^s$ – scene normalized fixated entity count.
  The effect of the visual scene on $c_e$ is considered.

### 7.1.3 User Activity Features (UA-Feat)

While interacting with the system, the user's activity can also be helpful in determining whether the user's eye gaze is tightly linked to the content of the speech. The following features are extracted from the user's activities:

- *maximal distance of the user's movements* – the maximal change of user position (3D coordinates) during the speech length.
  The user is expected to move within a smaller range while looking at entities and describing them.

- *variance of the user's positions*
  The user is expected to move less frequently while looking at entities and describing them.

### 7.1.4 Conversation Context Features (CC-Feat)

While talking to the system (i.e., the "expert"), the user's language and gaze behavior are influenced by the state of the conversation. For each speech-gaze instance, we use the previous system response type as a nominal feature to predict whether this is a closely coupled speech-gaze instance.

In our treasure hunting domain, there are eight types of system responses in two categories:
System-Initiative Responses:

- *specific-see* – the system asks whether the user sees a certain entity, e.g., "Do you see another couch?".

- *nonspecific-see* – the system asks whether the user sees anything, e.g., "Do you see anything else?", "Tell me what you see".

- *previous-see* – the system asks whether the user previously sees something, e.g., "Have you previously seen a similar object?".

- *describe* – the system asks the user to describe in detail what the user sees, e.g., "Describe it", "Tell me more about it".

- *compare* – the system asks the user to compare what the user sees, e.g., "Compare these objects".

- *clarify* – the system asks the user to make clarification, e.g., "I did not understand that", "Please repeat that".

- *action-request* – the system asks the user to take action, e.g., "Go back", "Try moving it".

User-Initiative Responses:

- *misc* – the system hands the initiative back to the user without specifying further requirements, e.g., "I don't know", "Yes".

## 7.2 Logistic Regression Model

Given the extracted feature $\mathbf{x}$ and the "closely coupled" label $y$ of each instance in the training set, we train a ridge logistic regression model (Cessie & Houwelingen, 1992) to predict whether an instance is a closely coupled instance ($y = 1$) or not ($y = 0$).

In the logistic regression model, the probability that $y_i = 1$, given the feature $\mathbf{x}_i = (x_1^i, x_2^i, \ldots, x_m^i)$, is modeled by

$$p(y_i|\mathbf{x}_i) = \frac{\exp(\sum_{j=1}^m \beta_j x_j^i)}{1 + \exp(\sum_{j=1}^m \beta_j x_j^i)}$$

where $\beta_j$ are the feature's weights to be learned.

The log-likelihood $l$ of the data $(\mathbf{X}, \mathbf{y})$ is

$$l(\beta) = \sum_i [y_i \log p(y_i|\mathbf{x}_i) + (1 - y_i) \log(1 - p(y_i|\mathbf{x_i}))]$$

In ridge logistic regression, parameters $\beta_j$ are estimated by maximizing a regularized log-likelihood

$$l^\lambda(\beta) = l(\beta) - \lambda ||\beta||^2$$

where $\lambda$ is the ridge parameter that is introduced to achieve more stable parameter estimation.

## 7.3 Evaluation of Speech-gaze Identification

Since the goal of identifying closely coupled speech-gaze instances is to improve word acquisition and we are only interested in acquiring nouns and adjectives, only the instances with recognized nouns/adjectives are used for training the logistic regression classifier. Among the 2969 instances with recognized nouns/adjectives and gaze fixations, 2002 (67.4%) instances are labeled as closely coupled. The speech-gaze prediction was evaluated by a 10-fold cross validation.

Table 1 shows the prediction precision and recall when different sets of features are used. As seen in the table, as more features are used, the prediction precision goes up and the recall goes down. It is important to note that prediction precision is more critical than recall for word acquisition when sufficient amount of data is available. *Noisy* instances where the gaze does not link to the speech content will only hurt word acquisition since they will guide the translation models to ground words to the wrong entities. Although higher recall can be helpful, its effect is expected to become less when co-occurrences can already be established.

The results show that speech features (S-Feat) and conversation context features (CC-Feat), when used alone, do not improve prediction precision much compared to the baseline of predicting all instances "closely coupled" with a precision of 67.4%. When used alone, gaze features (G-Feat) and user activity features (UA-Feat) are the two most useful feature sets for increasing prediction precision. When they are used together, the prediction precision is further increased. Adding either speech features or conversation context features to gaze and user activity features (G-Feat + UA-Feat + S-Feat/CC-Feat) increases the prediction precision more. Using all four sets of features (G-Feat + UA-Feat + S-Feat +

| Feature sets | Precision | Recall |
|:---:|:---:|:---:|
| Null (*baseline*) | 0.674 | 1 |
| S-Feat | 0.686 | 0.995 |
| G-Feat | 0.707 | 0.958 |
| UA-Feat | 0.704 | 0.942 |
| CC-Feat | 0.688 | 0.936 |
| G-Feat + UA-Feat | 0.719 | 0.948 |
| G-Feat + UA-Feat + S-Feat | 0.741 | 0.908 |
| G-Feat + UA-Feat + CC-Feat | 0.731 | 0.918 |
| G-Feat + UA-Feat + S-Feat + CC-Feat | **0.748** | 0.899 |

Table 1: Speech-gaze prediction performances with different feature sets

CC-Feat) achieves the highest prediction precision. McNemar tests have shown that this is a significant change compared to using G-Feat + UA-Feat + S-Feat ($\chi^2 = 8.3, p < 0.004$) and all other feature configurations ($\chi^2 = 45.4 \sim 442.7, p < 0.0001$). Therefore, we choose to use all feature sets to identify the closely coupled speech-gaze instances for word acquisition.

## 8. Evaluation of Word Acquisition

Each practical conversational system starts with an initial knowledge base (vocabulary). We assume that the system already has one default word for each entity in its *default vocabulary*. The default word of an entity indicates the semantic type of the entity. For example, the word "barrel" is the default word for the entity *barrel*. Among the acquired words, we only evaluate those new words that are not in the system's vocabulary. For example, the word "barrel" would be excluded from the candidate words acquired for the entity *barrel*.

### 8.1 Grounding Words to Domain Concepts

Based on the translation models for word acquisition (Sections 5 & 6), we can obtain the word-entity association probability $p(w|e)$. This probability provides a means to ground words to entities. In conversational systems, one important goal of word acquisition is to make the system understand the semantic meaning of new words. Word acquisition by grounding words to objects is not always sufficient for identifying their semantic meanings. Suppose the word *green* is grounded to a green chair object, so is the word *chair*. Although the system is aware that *green* is some word describing the green chair, it does not know that the word *green* refers to the chair's color while the word *chair* refers to the chair's semantic type. Thus, after learning the word-entity associations $p(w|e)$ by the translation models, we need to further ground words to domain concepts of entity properties.

Based on the domain model discussed earlier (Section 6.1), we apply WordNet to ground words to domain concepts. For each entity $e$, based on association probabilities $p(w|e)$, we can choose the $n$-best words as acquired words for $e$. Those $n$-best words have the $n$ highest association probabilities. For each word $w$ acquired for $e$, the grounded concept $c_e^*$ for $w$ is chosen as the one that has the highest semantic relatedness with $w$:

$$c_e^* = \arg\max_i [\max_j sim(s(c_e^i), s_j(w))] \tag{12}$$

where $sim(s(c_e^i), s_j(w))$ is the semantic similarity score defined in Equation (6).

To evaluate the acquired words for the domain concepts, we manually compile a set of "gold standard" words from all users' speech transcripts and gaze fixations. Those "gold standard" words are the words that the users have used to refer to the entities and their properties (e.g., color, size, shape) during the interaction with the system. The automatically acquired words are evaluated against those "gold standard" words.

## 8.2 Evaluation Metrics

Following the standard evaluation on information retrieval, the following metrics are used to evaluate the words acquired for domain concepts (i.e., entity properties) $\{c_e\}$.

- Precision

$$\frac{\sum_{c_e} \# \text{ words correctly acquired for } c_e}{\sum_{c_e} \# \text{ words acquired for } c_e}$$

- Recall

$$\frac{\sum_{c_e} \# \text{ words correctly acquired for } c_e}{\sum_{c_e} \# \text{ "gold standard" words of } c_e}$$

- Mean Average Precision (MAP)

$$\text{MAP} = \frac{\sum_e \dfrac{\sum_{r=1}^{N_w} P(r) \times rel(r)}{N_e}}{\#e}$$

where $N_e$ is the number of the "gold standard" words of all the properties $c_e$ of entity $e$, $N_w$ is the vocabulary size, $P(r)$ is the acquisition precision at a given cut-off rank $r$, $rel(r)$ is a binary function indicating whether the word at rank $r$ is a "gold-standard" word for some property $c_e$ of entity $e$.

## 8.3 Evaluation Results

To investigate the effects of speech-gaze temporal information and domain semantic knowledge on word acquisition, we compare the word acquisition performances of the following models:

- Model-1 – base model I without word-entity alignment (Equation (1)).

- Model-1-r – Model-1 with semantic relatedness rescoring of word-entity association.

- Model-2 – base model II with positional alignment (Equation (2)).

- Model-2s – enhanced model with semantic alignment (Equation (8)).

- Model-2t – enhanced model with temporal alignment (Equation (3)).

- Model-2ts – enhanced model with both temporal and semantic alignment (Equation (10)).

- Model-2t-r – Model-2t with semantic relatedness rescoring of word-entity association.

### 8.3.1 RESULTS OF USING SPEECH-GAZE TEMPORAL INFORMATION

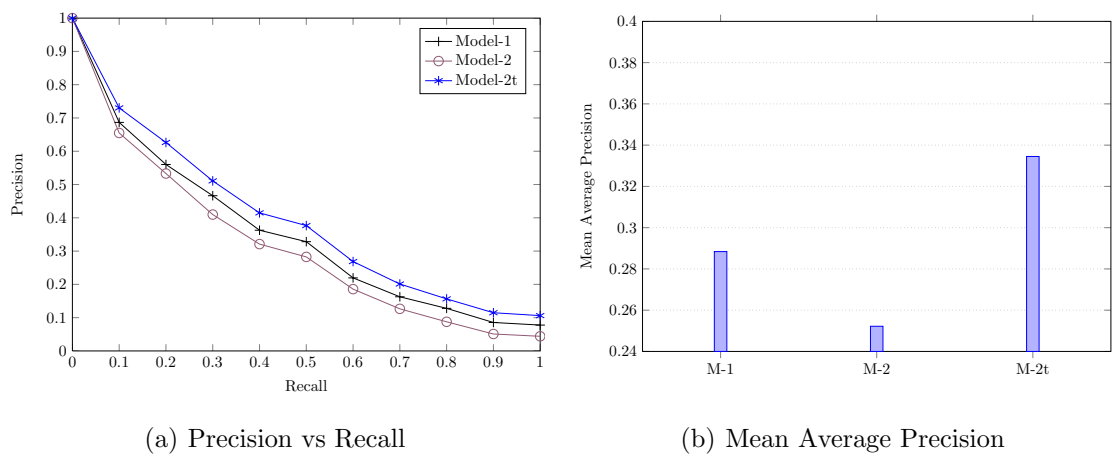

(a) Precision vs Recall          (b) Mean Average Precision

Figure 5: Word acquisition performance when speech-gaze temporal information is used

Figure 5 shows the interpolated precision-recall curves and Mean Average Precisions (MAPs) of Model-2t and the baseline models Model-1 and Model-2. As shown in the figure, Model-2 does not improve word acquisition compared to Model-1. This result shows that it is not helpful to consider the index-based positional alignment of word and entity for word acquisition. By incorporating speech-gaze temporal alignment, Model-2t consistently achieves higher precisions than Model-1 at different recalls. In terms of MAP, Model-2t significantly increases MAP ($t = 3.08, p < 0.002$) compared to Model-1. This means that the use of speech-gaze temporal alignment improves word acquisition.

### 8.3.2 RESULTS OF USING DOMAIN SEMANTIC RELATEDNESS

Figure 6 shows the results of using domain semantic relatedness in word acquisition. As shown in the figure, compared to the baseline of using no extra knowledge (Model-1), using domain semantic relatedness improves word acquisition no matter when it is used to rescore word-entity association (Model-1-r) or to constrain word-entity alignment (Model-2s). Compared to Model-1, the MAP is significantly improved by Model-1-r ($t = 6.32, p < 0.001$) and Model-2s ($t = 5.36, p < 0.001$).

Domain semantic relatedness can also be used together with speech-gaze temporal information to improve word acquisition. Compared to Model-1, the MAP is significantly increased by Model-2ts ($t = 5.59, p < 0.001$) that uses semantic relatedness together with temporal information to constrain word-entity alignments and Model-2t-r ($t = 6.01, p < 0.001$), where semantic relatedness is used to rescore the word-entity associations learned by Model-2t.

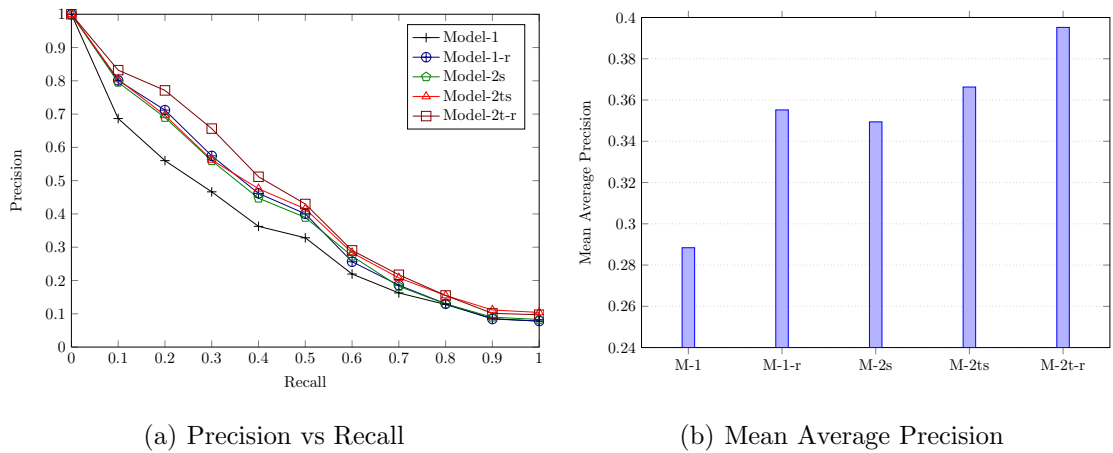

(a) Precision vs Recall     (b) Mean Average Precision

Figure 6: Word acquisition performance when domain semantic relatedness is used

Comparing the two ways of using semantic relatedness for word acquisition, it is found that rescoring word-entity association with semantic relatedness works better. Model-2t-r achieve higher MAP ($t = 2.22, p < 0.015$) than Model-2ts.

It is also verified that using both speech-gaze temporal alignment and domain semantic relatedness rescoring works better than using either one alone. With temporal alignment and semantic relatedness rescoring, Model-2t-r significantly increases MAP when compared to Model-1-r ($t = 2.75, p < 0.004$) where only semantic relatedness rescoring is used and Model-2t ($t = 5.38, p < 0.001$) where only temporal alignment is used.

### 8.3.3 Results Based on Identified Closely Coupled Speech-Gaze Streams

We have shown that Model-2t-r, where both speech-gaze temporal alignment and domain semantic relatedness rescoring are incorporated, achieves the best word acquisition performance. Therefore, Model-2t-r is used to evaluate the word acquisition based on the identified closely coupled speech-gaze data. Since Model-2t-r requires linking domain models with external knowledge source (e.g., WordNet) which may not be available for some applications, we also evaluate the effect of the identification of closely coupled speech-gaze streams on word acquisition with Model-2t, where only speech-gaze temporal alignment is incorporated.

We evaluate the effect of automatic identification of closely coupled speech-gaze instances on word acquisition through a 10-fold cross validation. In each fold, 10% of the data set was used to train the logistic regression classifier for predicting closely coupled speech-gaze instances, then all instances, predicted closely coupled instances, and true (manually labeled) closely coupled instances of the other 90% of the data set were used for word acquisition respectively. Figures 7 & 8 show the averaged interpolated precision-recall curves and MAPs achieved by Model-2t and Model-2t-r using all instances (*all*), only predicted closely coupled instances (*predicted*), and true closely coupled instances (*true*).

When words are acquired by Model-2t, as shown in Figure 7, using predicted closely coupled instances achieves better performance than using all instances. The MAP is significantly increased ($t = 2.69, p < 0.005$) by acquiring words from predicted closely coupled instances. The result shows that the identification of closely coupled speech-gaze instances helps word acquisition. When the true closely coupled speech-gaze instances are used for word acquisition, the acquisition performance is further improved. This means that better identification of closely coupled speech-gaze instances can lead to better word acquisition performance.

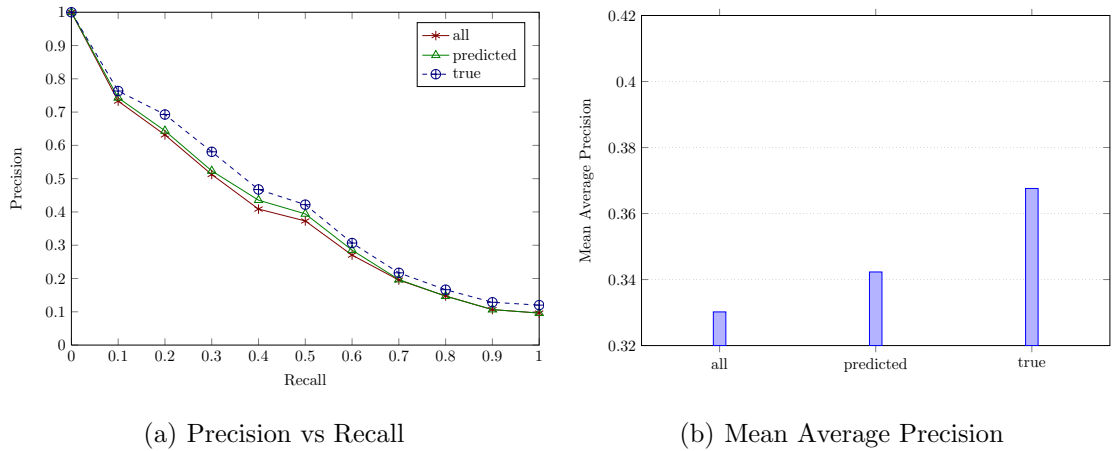

(a) Precision vs Recall

(b) Mean Average Precision

Figure 7: Word acquisition performance by Model-2t on different data set

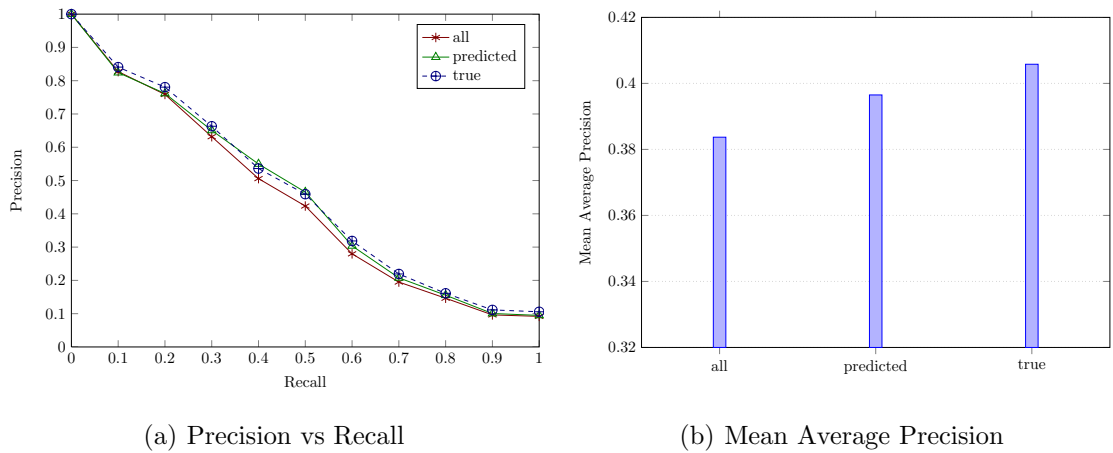

(a) Precision vs Recall

(b) Mean Average Precision

Figure 8: Word acquisition performance by Model-2t-r on different data set

When words are acquired by Model-2t-r, as shown in Figure 8, using predicted closely coupled instances improves acquisition performance compared to using all instances. By acquiring words from predicted closely coupled speech-gaze instances, the MAP is increased ($t = 1.81, p < 0.037$) although this improvement is less significant than the one with Model-2t.

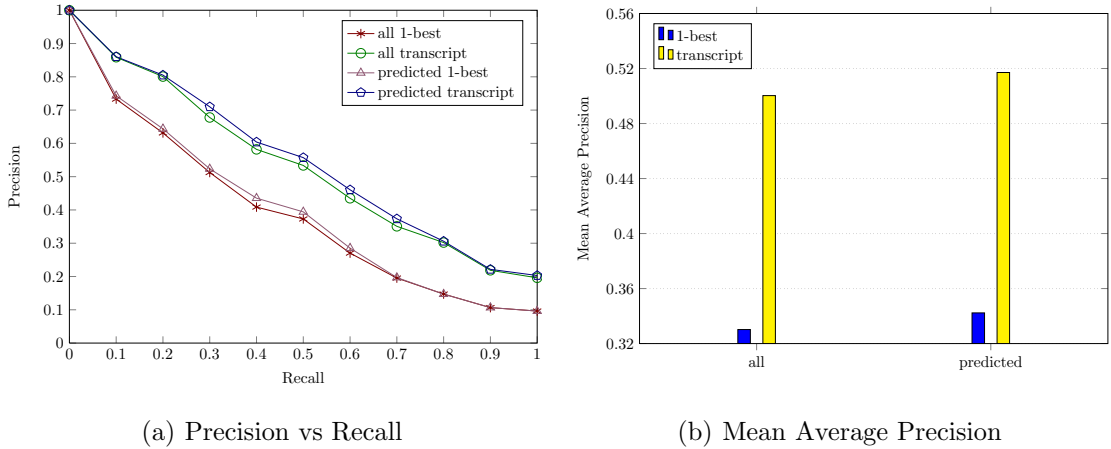

(a) Precision vs Recall                    (b) Mean Average Precision

Figure 9: Word acquisition performance by Model-2t on speech recognition and transcript

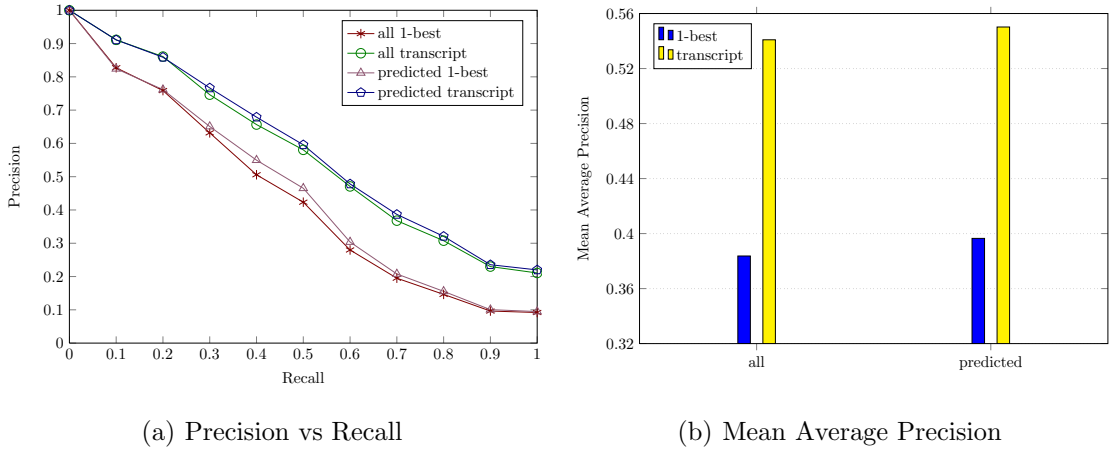

(a) Precision vs Recall                    (b) Mean Average Precision

Figure 10: Word acquisition performance by Model-2t-r on speech recognition and transcript

### 8.3.4 Comparison of Results Based on Speech Recognition and Transcript

To show the effect of speech recognition quality on word acquisition, we also compare the acquisition performances based on speech transcript and the 1-best recognition. When word acquisition is based on speech transcript, the word sequence in the parallel speech-gaze data set contains nouns and adjectives in the speech transcript. Accordingly, the speech feature used for coupled speech-gaze identification is extracted from the speech transcript.

Figures 9 & 10 show the word acquisition performances of Model-2t and Model-2t-r using all instances and using only predicted coupled instances based on speech transcript and the 1-best recognition respectively. As shown in the figures, the quality of speech recognition is critical to word acquisition performance. As expected, word acquisition performance based on speech transcript is much better than on recognized speech.

## 9. Examples

Table 2 shows the 10-best candidate words acquired for the entity *couch* by Model-1, Model-2t, and Model-2t-r based on all speech-gaze instances and Model-2t-r based on predicted closely coupled instances. The probabilities of these candidate words are also given in the table. Across all these models, although the same four words (shown in bold font) are acquired by each model, the ranking of the acquired words achieves the best by Model-2t-r based on predicted closely coupled instances.

Table 3 shows another example of the 10-best candidate words acquired for the entity *stool* by the four different models. Model-1 acquires four correct words in the 10-best list. Although Model-2t also acquires four correct words in the 10-best list, the rankings of these words are higher. With both speech-gaze temporal alignment and domain semantic relatedness rescoring, Model-2t-r acquires seven correct words in the 10-best list. The rankings of these correct words are also improved. Compared to using all instances with Model-2t-r, although using the predicted coupled instances with Model-2t-r results in the same seven correct words with the same ranks in the 10-best list, the probabilities of these correctly acquired words are higher. This means that the results based on the predicted coupled instances are more confident.

| Model | Model-1 | Model-2t | Model-2t-r | Model-2t-r(predicted) |
|---|---|---|---|---|
| Rank 1 | **couch**(0.1105) | **couch**(0.1224) | **couch**(0.4743) | **couch**(0.4667) |
| Rank 2 | bedroom(0.1047) | **chair**(0.0798) | **chair**(0.1668) | **chair**(0.1557) |
| Rank 3 | **chair**(0.1004) | bed(0.0593) | **bench**(0.0949) | **bench**(0.11129) |
| Rank 4 | bad(0.0936) | small(0.0536) | bed(0.0311) | bed(0.0368) |
| Rank 5 | room(0.0539) | room(0.0528) | small(0.0235) | small(0.0278) |
| Rank 6 | wooden(0.0354) | bad(0.0489) | bad(0.0226) | bad(0.0265) |
| Rank 7 | **bench**(0.0319) | **yellow**(0.0333) | room(0.0174) | room(0.0137) |
| Rank 8 | small(0.0289) | **bench**(0.0332) | lot(0.0151) | **yellow**(0.0127) |
| Rank 9 | **yellow**(0.0274) | lot(0.0331) | **yellow**(0.0107) | couple(0.0101) |
| Rank 10 | couple(0.0270) | wooden(0.0226) | couple(0.0085) | lot(0.0090) |

Table 2: N-best candidate words acquired for the entity *couch* by different models

| Model | Model-1 | Model-2t | Model-2t-r | Model-2t-r(predicted) |
|---|---|---|---|---|
| Rank 1 | plant(0.0793) | plant(0.0592) | **stool**(0.1457) | **stool**(0.1532) |
| Rank 2 | room(0.0508) | room(0.0440) | **little**(0.1435) | **little**(0.1509) |
| Rank 3 | **little**(0.0471) | **little**(0.0410) | **small**(0.1412) | **small**(0.1490) |
| Rank 4 | flower(0.0424) | flower(0.0409) | **footstool**(0.0573) | **footstool**(0.0602) |
| Rank 5 | stairs(0.0320) | **square**(0.0408) | **ottoman**(0.0572) | **ottoman**(0.0601) |
| Rank 6 | call(0.0319) | **small**(0.0403) | ground(0.0275) | ground(0.0289) |
| Rank 7 | **square**(0.0302) | next(0.0308) | media(0.0263) | media(0.0276) |
| Rank 8 | **footstool**(0.0301) | **stool**(0.0307) | **chair**(0.0257) | **chair**(0.0272) |
| Rank 9 | brown(0.0300) | brown(0.0300) | plant(0.0253) | plant(0.0270) |
| Rank 10 | **short**(0.0294) | stairs(0.0226) | **square**(0.0234) | **square**(0.0247) |

Table 3: N-best candidate words acquired for the entity *stool* by different models

## 10. The Effect of Online Word Acquisition on Language Understanding

One important goal of word acquisition is to use the acquired new words to help language understanding in subsequent conversation. To demonstrate the effect of online word acquisition on language understanding, we conduct simulation studies based on our collected data. In these simulations, the system starts with an initial knowledge base – a vocabulary of words associated to domain concepts. The system continuously enhances its knowledge base by acquiring words from users with Model-2t-r that incorporates both speech-gaze temporal information and domain semantic relatedness. The enhanced knowledge base is used to understand the language of new users.

We evaluate language understanding performance on concept identification rate (CIR):

$$CIR = \frac{\#\text{correctly identified concepts in the 1-best speech recognition}}{\#\text{concepts in the speech transcript}}$$

We simulate the process of online word acquisition and evaluate its effect on language understanding for two situations: 1) the system starts with no training data but with a small initial vocabulary, and 2) the system starts with some training data.

### 10.1 Simulation 1: When the System Starts with No Training Data

To build conversational systems, one approach is that domain experts provide domain vocabulary to the system at design time. Our first simulation follows this practice. The system is provided with a default vocabulary to start without training data. The default vocabulary contains one "seed" word for each domain concept.

Using the collected data of 20 users, the simulation process goes through the following steps:

- For user index $i = 1, 2, \ldots, 20$:

    - Evaluate CIR of the $i$-th user's utterances (1-best speech recognition) with the current system vocabulary.
    - Acquire words from all the instances (with 1-best speech recognition) of users $1 \cdots i$.
    - Among the 10-best acquired words, add verified new words to the system vocabulary.

In the above process, the language understanding performance on each individual user depends on the user's own language as well as the user's position in the user sequence. To reduce the effect of user ordering on language understanding performance, the above simulation process is repeated 1000 times with randomly ordered users. The average of the CIRs in these simulations is shown in Figure 11.

Figure 11 also shows the CIRs when the system is with a static knowledge base (vocabulary). The curve is drawn in the same way as the curve with a dynamic knowledge base, except without word acquisition in the random simulation processes. As we can see in the figure, when the system doest not have word acquisition capability, its language understanding performance does not change after more users have communicated to the system. With the capability of automatic word acquisition, the system's language understanding performance becomes better after more users have talked to the system.

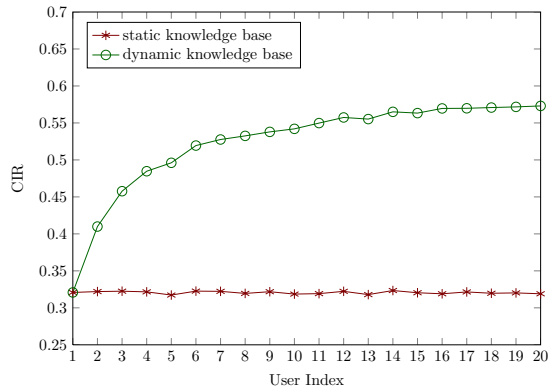

Figure 11: CIR of user language achieved by the system starting with no training data

## 10.2 Simulation 2: When the System Starts with Training Data

Many conversational systems use real user data to derive domain vocabulary. To follow this practice, the second simulation provides the system with some training data. The training data serves two purposes: 1) build an initial vocabulary of the system; 2) train a classifier to predict the closely coupled speech-gaze instances of new users' data.

Using the collected data of 20 users, the simulation process goes through the following steps:

- Using the first $m$ users' data as training data, acquire words from the training instances (with speech transcript); add the verified 10-best words to the system's vocabulary as "seed" words; build a classifier with the training data for prediction of closely coupled speech-gaze instances.

- Evaluate the effect of incremental word acquisition on CIR of the remaining (20-$m$) users' data. For user index $i = 1, 2, \ldots, (20$-$m)$:

  - Evaluate CIR of the $i$-th user's utterances (1-best speech recognition).
  - Predict closely coupled speech-gaze instances of the $i$-th user's data.
  - Acquire words from the $m$ training users' true coupled instances (with speech transcript) and the predicted coupled instances (with 1-best speech recognition) of users $1 \cdots i$.
  - Among the 10-best acquired words, add verified new words to the system vocabulary.

The above simulation process is repeated 1000 times with randomly ordered users to reduce the effect of user ordering on the language understanding performance. Figure 12 shows the averaged language understanding performance of these random simulations.

The language understanding performance of the system with a static knowledge base is also shown in Figure 12. The curve is drawn by the same random simulations without the steps of word acquisition. We can observe a general trend in the figure that, with word acquisition, the system's language understanding becomes better after more users have communicated to the system. Without word acquisition capability, the system's language understanding performance does not increase after more users have conversed with the system.

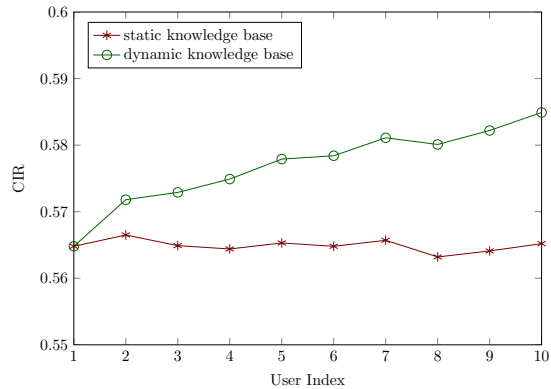

Figure 12: CIR of user language achieved by the system starting with training data of 10 users

The simulations show that automatic vocabulary acquisition is beneficial to the system's language understanding performance when training data is available. When training data is not available, vocabulary acquisition could be more important and beneficial to robust language understanding.

### 10.3 The Effect of Speech Recognition on Online Word Acquisition and Language Understanding

The simulation results in Figures 11 & 12 are based on the 1-best recognized speech hypotheses with a relatively high WER (48.1%). With better speech recognition, the system will have better concept identification performance. To show the effect of speech recognition quality on online word acquisition and language understanding, we also perform Simulation 1 and Simulation 2 based on speech transcript. The simulation processes are the same as the ones based on the 1-best speech recognition except that word acquisition is based on speech transcript and CIR is evaluated also on speech transcript in the new simulations.

Figure 13 shows the CIR curves based on speech transcript during online conversation. With word acquisition, the system's language understanding becomes better after more users have communicated to the system. This is consistent with the CIR curves based on the 1-best speech recognition. However, the CIRs based on speech transcript is much higher the CIRs based on the 1-best speech recognition, which verifies that speech recognition quality is critical to language understanding performance.

## 11. Discussion and Future Work

Our experimental results have shown that incorporating extra information improves word acquisition compared to completely unsupervised approaches. However, our current ap-

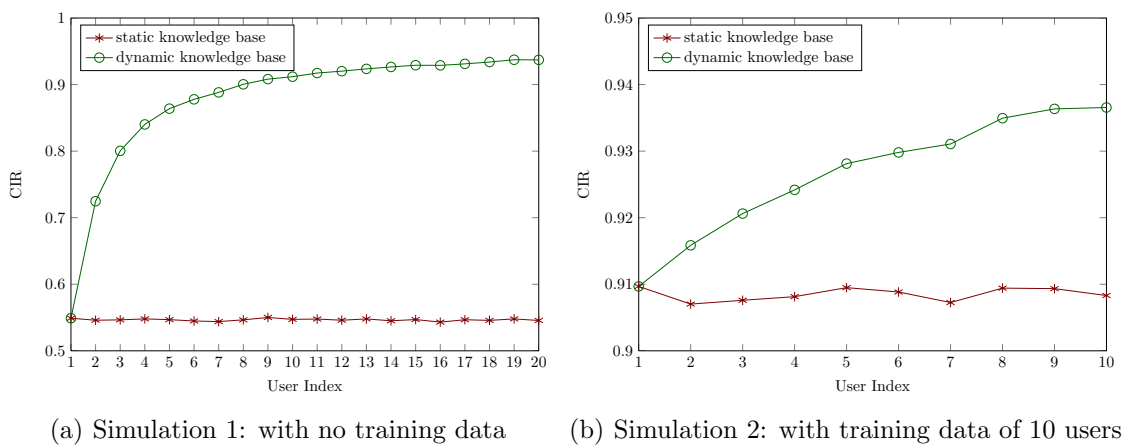

(a) Simulation 1: with no training data       (b) Simulation 2: with training data of 10 users

Figure 13: CIR of user language (transcript) achieved by the system during online conversation

proaches have several limitations. First, the incorporation of domain knowledge through semantic relatedness based on WordNet will restrict the acquired words to those appear in WordNet. This is certainly not desirable. But this limitation can be readily addressed by changing the way how the word probability distribution is tailored by semantic relatedness (in Section 6.3.1 and Section 6.3.2). For example, one simple way is to keep the probability mass for those words not in WordNet and only tailor the distribution from those words that occur in WordNet based on their semantic relatedness with the object.

Second, in our current approach, acquired words are limited to the words that are recognized by the speech recognizer. As shown in Section 8.3.4, the speech recognition performance is rather poor in our experiments. This is partly due to the lack of language models specifically trained for this domain. Approaches to improve speech recognition, for example, based on a referential semantic language model described in (Schuler, Wu, & Schwartz, 2009) will potentially improve acquisition performance. Furthermore, the set of acquired words is bounded by the vocabulary of the speech recognizer. Any new words that are not in the dictionary will not be acquired. To break this barrier, inspired by previous work (Yu & Ballard, 2004; Taguchi et al., 2009), we are currently extending our approach to incorporate grounding acoustic phoneme sequences to domain concepts.

Another limitation of our current approaches is that they are incapable of acquiring multiword expressions. They can only map single words to domain concepts. However, we did observe multiword expressions (e.g., Rubik's cube) in our data. We will examine this issue in our future work by incorporating more linguistic knowledge and by modeling "fertility" of entities, for example, as in IBM Model 3 and 4.

The simplicity of our current models also limits word acquisition performance. For example, the alignment model based on temporal information directly incorporates findings from psycholinguistic studies. These studies were generally conducted in a much simpler settings without interaction. The recent work by Fang, Chai, and Ferreira (2009) has shown correlations between intensity of gaze fixations and objects denoted by linguistic centers (e.g., forward-looking centers based on centering theory, Grosz, Joshi, & Weinstein, 1995). We plan to incorporate these results to improve alignment modeling in the future.

To further improve performance, another interesting direction is to take into consideration of the interactive nature of conversation, for example, by combining dialog management to solicit user feedback on the acquired words. However, it will be important to identify strategies to balance the trade off between explicit feedback solicitation (and thus lengthening the interaction) and the quality of the acquired words. Reinforcement learning can be a potential approach to address this problem.

## 12. Conclusions

Motivated by the psycholinguistic findings, we investigate the use of eye gaze for automatic word acquisition in multimodal conversational systems. This paper presents several enhanced models that incorporate user language behavior, domain knowledge, and conversation context in word acquisition. Our experiments have shown that these enhanced models significantly improve word acquisition performance.

Recent advancement in eye tracking technology has made available non-intrusive eye tracking devices that can tolerate head motion and provide high tracking quality. Integrating eye tracking with conversational interfaces is no longer beyond reach. We believe that incorporating eye gaze with automatic word acquisition provides another potential approach to improve the robustness of human-machine conversation.

## Acknowledgments

This work was supported by IIS-0347548 and IIS-0535112 from the National Science Foundation. We would like to thank anonymous reviewers for their valuable comments and suggestions.